\documentclass{llncs}

\usepackage{amssymb,amsfonts,mathrsfs,bm}
\usepackage{graphicx}
\usepackage{subfigure}

\usepackage{url}
\urldef{\mailsa}\path|{alfred.hofmann, ursula.barth, ingrid.beyer, christine.guenther,|
\urldef{\mailsb}\path|frank.holzwarth, anna.kramer, erika.siebert-cole, lncs}@springer.com|
\newcommand{\keywords}[1]{\par\addvspace\baselineskip
\noindent\keywordname\enspace\ignorespaces#1}

\begin{document}

\mainmatter  

\title{Multi-view Laplacian Support Vector Machines}

\titlerunning{Multi-view Laplacian SVMs}

%
%
\author{Shiliang Sun}
\authorrunning{S. Sun}

\institute{Department of Computer Science and Technology, \\East
China Normal University, Shanghai 200241,
China\\
\email{slsun@cs.ecnu.edu.cn}
}

%
%

\toctitle{Lecture Notes in Computer Science}
\tocauthor{Authors' Instructions}
\maketitle

\begin{abstract}
We propose a new approach, multi-view Laplacian support vector
machines (SVMs), for semi-supervised learning under the multi-view
scenario. It integrates manifold regularization and multi-view
regularization into the usual formulation of SVMs and is a natural
extension of SVMs from supervised learning to multi-view
semi-supervised learning. The function optimization problem in a
reproducing kernel Hilbert space is converted to an optimization in
a finite-dimensional Euclidean space. After providing a theoretical
bound for the generalization performance of the proposed method, we
further give a formulation of the empirical Rademacher complexity
which affects the bound significantly. From this bound and the
empirical Rademacher complexity, we can gain insights into the roles
played by different regularization terms to the generalization
performance. Experimental results on synthetic and real-world data
sets are presented, which validate the effectiveness of the proposed
multi-view Laplacian SVMs approach. \keywords{graph Laplacian,
multi-view learning, reproducing kernel Hilbert space,
semi-supervised learning, support vector machine}
\end{abstract}

\section{Introduction}

Semi-supervised learning or learning from labeled and unlabeled
examples has attracted considerable attention in the last
decade~\cite{Chapelle06SSLbook,Culp08GraBasSSL,Sun10JMLR}. This is
partially motivated by the fact that for many practical applications
collecting a large number of unlabeled data is much less involved
than collecting labeled data considering the expensive and tedious
annotation process. Moreover, as human learning often occurs in the
semi-supervised learning manner (for example, children may hear some
words but do not know their exact meanings), research on
semi-supervised learning also has the potential to uncover insights
into mechanisms of human learning~\cite{Zhu08tecrep}.

In some machine learning applications, examples can be described by
different kinds of information. For example, in television broadcast
understanding, broadcast segments can be simultaneously described by
their video signals and audio signals which can be regarded as
information from different properties or different ``views".
Multi-view semi-supervised learning, the focus of this paper,
attempts to perform inductive learning under such circumstances.
However, it should be noted that if there are no natural multiple
views, artificially generated multiple views can still work
favorably~\cite{Sun11VC}.

In this paper we are particularly interested in multi-view
semi-supervised learning approaches derived from support vector
machines (SVMs)~\cite{VapnikSLT98}. As a state-of-the-art method in
machine learning, SVMs not only are theoretically well justified but
also show very good performance for real applications. The
transductive SVMs~\cite{Joachims99TSVM},
S$^3$VMs~\cite{Bennett99S3VM,Fung01S3VM} and Laplacian
SVMs~\cite{Belkin06ManfReg} have been proposed as extensions of SVMs
from supervised learning to single-view semi-supervised learning.
For multi-view learning, there are also several extensions of SVMs
proposed such as the co-Laplacian SVMs~\cite{Sindhwani05CoLapSVM}
and SVM-2K~\cite{Farquhar06svm2k}.

Regularization theory is an important technique in mathematics and
machine learning~\cite{Tikhonov63reg,Evgeniou00RegSVM}. Many methods
can be explained from the point of view of regularization. A close
parallel to regularization theory is capacity control of function
classes~\cite{JohnbookKerMat04}. Both regularization and capacity
control of function classes can play a central role in alleviating
over-fitting of machine learning algorithms.

The new method, multi-view Laplacian SVMs, proposed in this paper
can also be explained by regularization theory and capacity control
of function classes. It integrates three regularization terms
respectively on function norm, manifold and multi-view
regularization. As an appropriate integration of them and thus the
effective use of information from labeled and unlabeled data, our
method has the potential to outperform many related counterparts.
The different roles of these regularization terms on capacity
control will be unfolded later as a result of our empirical
Rademacher complexity analysis. Besides giving the bound on the
generalization error, we also report experimental results of the
proposed method on synthetic and real-world data sets.

The layout of this paper is as follows. Section~\ref{secMvLap}
introduces the objective function of the proposed approach with
concerns on different regularization terms, and its optimization.
Theoretical insights on the generalization error and the empirical
Rademacher complexity are  covered by Section~\ref{secTheAna}. Then,
experimental results are reported in Section~\ref{secExp}. Finally,
conclusions are drawn in Section~\ref{secCon}.

\section{Multi-view Laplacian SVMs (MvLapSVM)}
\label{secMvLap}

\subsection{Manifold Regularization}
Let $x_1,\ldots,x_{l+u} \in \textbf{R}^d$ denote a set of inputs
including $l$ labeled examples and $u$ unlabeled ones with label
space $\{+1,-1\}$. For manifold regularization, a data adjacency
graph $W_{(l+u)\times(l+u)}$ is defined whose entries measure the
similarity or closeness of every pair of inputs. We use a typical
construction of $W$: $W_{ij}=0$ for most pairs of inputs, and for
neighboring $x_i,x_j$ the corresponding entry is given by
\begin{equation}
W_{ij}=\exp(-\|x_i-x_j\|^2/2\sigma^2),
\end{equation}
where $\|x_i-x_j\|$ is the Euclidean norm in $\textbf{R}^d$.

The manifold regularization functional acting on any function $f:
\textbf{R}^d\rightarrow \textbf{R}$ is defined as
follows~\cite{Belkin03LapEig}
\begin{equation}
M_{reg}(f)=\frac{1}{2}\sum_{i,j=1}^{l+u}W_{ij}(f(x_i)-f(x_j))^2.
\end{equation}
It is clear that a smaller $M_{reg}(f)$ indicates a smoother
function $f$. Define vector
$\textbf{f}=(f(x_1),\ldots,f(x_{l+u}))^\top$. Then
\begin{eqnarray}
M_{reg}(f)&=&\sum_{i=1}^{l+u}(\sum_{j=1}^{l+u}W_{ij})f^2(x_i)-\sum_{i,j=1}^{l+u}W_{ij}f(x_i)f(x_j)\nonumber\\
    &=&\textbf{f}^\top(V-W)\textbf{f},
\end{eqnarray}
where matrix $V$ is diagonal with the $i$th diagonal entry
$V_{ii}=\sum_{j=1}^{l+u}W_{ij}$. The matrix $L\triangleq V-W$, which
is arguably positive semidefinite, is called the graph Laplacian of
$W$. In our empirical studies in Section~\ref{secExp}, a normalized
Laplacian $\bar{L}=V^{-1/2}LV^{-1/2}$ is used because this
normalized one often performs as well or better in practical
tasks~\cite{Belkin06ManfReg}.

\subsection{Multi-view Regularization}

For multi-view learning, an input $x\in \textbf{R}^d$ can be
decomposed into components corresponding to multiple views, such as
$x=(x^1,\ldots, x^m)$ for an $m$-view representation. A function
$f_j$ defined on view $j$ only depends on $x^j$, while ignoring the
other components $(x^1,\ldots, x^{j-1},x^{j+1},\dots,x^m)$.

For multi-view semi-supervised learning, there is a commonly
acceptable assumption that a good learner can be learned from each
view~\cite{Blum98Cot}. Consequently, these good learners in
different views should be consistent to a large extent with respect
to their predictions on the same examples. We also adopt this
assumption and use the regularization idea to wipe off those
inconsistent learners. Given the $l+u$ examples, the multi-view
regularization functional for $m$ functions $f_1,\ldots, f_m$ can be
formulated as
\begin{equation}
V_{reg}(f_1,\ldots,
f_m)=\sum_{j>k,k=1}^{m}\sum_{i=1}^{l+u}[f_{j}(x_{i})-f_{k}(x_{i})]^{2}.
\end{equation}
Clearly, a smaller $V_{reg}(f_1,\ldots, f_m)$ tends to find good
learners in each view.

\subsection{MvLapSVM}
As is usually assumed in multi-view learning, each view is regarded
to be sufficient to train a good learner. Therefore, we can write
the final prediction as $f=\frac{1}{m} \sum_{i=1}^m f_i$. For
MvLapSVM, in this paper we concentrate on the two-view case, that is
$m=2$. In this scenario, the objective function for MvLapSVM is
defined as
\begin{eqnarray}
\label{eqnTwoViews} \min_{f_1\in \mathcal {H}_1, f_2\in \mathcal
{H}_2}
&&\frac{1}{2l}\sum_{i=1}^{l}[(1-y_{i}f_1(x_{i}))_{+}+(1-y_{i}f_2(x_{i}))_{+}]+\nonumber\\
&&\gamma_1(\|f_1\|^{2}+\|f_2\|^{2})+\frac{\gamma_{2}}{(l+u)^2}({\textbf{f}_1}^{\top}L_1{\textbf{f}_1}+\nonumber \\
&&{\textbf{f}_2}^{\top}L_2{\textbf{f}_2})+
\frac{\gamma_{3}}{(l+u)}\sum_{i=1}^{l+u}[f_{1}(x_{i})-f_{2}(x_{i})]^{2},
\end{eqnarray}
where $\mathcal {H}_1, \mathcal {H}_2$ are the reproducing kernel
Hilbert spaces~\cite{Aronszajn50,Sindhwani08anRKHS} in which $f_1,
f_2$ are defined, nonnegative scalars $\gamma_1, \gamma_2, \gamma_3$
are respectively norm regularization, manifold regularization and
multi-view regularization coefficients, and vector
$\textbf{f}_1=(f_1(x_1),...,f_1(x_{l+u}))^\top$,
$\textbf{f}_2=(f_2(x_1),...,f_2(x_{l+u}))^\top$.

\subsection{Optimization}
We now concentrate on solving
(\ref{eqnTwoViews}). As an application of the representer
theorem~\cite{Kimeldorf71,RosenbergDiss08},  the solution to problem
(\ref{eqnTwoViews}) has the following form
\begin{equation}
\label{eqn041702} f_1(x)=\sum_{i=1}^{l+u}\alpha_{1}^{i}k_1(x_{i},x),
\; f_2(x)=\sum_{i=1}^{l+u}\alpha_{2}^{i}k_2(x_{i},x).
\end{equation}
Therefore, we can rewrite $\|f_1\|^{2}$ and $\|f_2\|^{2}$ as
\begin{equation}
\|f_{1}\|^2=\boldsymbol{\alpha}_1^\top K_1 \boldsymbol{\alpha}_1, \;
\|f_{2}\|^2=\boldsymbol{\alpha}_2^\top K_2 \boldsymbol{\alpha}_2,
\end{equation}
where $K_1$ and $K_2$ are $(l+u)\times(l+u)$ Gram matrices
respective from view $\mathcal {V}^1$ and $\mathcal {V}^2$, and
vector
$\boldsymbol{\alpha}_1=(\alpha_{1}^{1},...,\alpha_{1}^{l+u})^\top$,
$\boldsymbol{\alpha}_2=(\alpha_{2}^{1},...,\alpha_{2}^{l+u})^\top$.
In addition, we have
\begin{equation}
\textbf{f}_1=K_1\boldsymbol{\alpha}_1, \;
\textbf{f}_2=K_2\boldsymbol{\alpha}_2.
\end{equation}

To simplify our formulations, we respectively replace
$\frac{\displaystyle \gamma_{2}}{\displaystyle (l+u)^2}$ and
$\frac{\displaystyle \gamma_{3}}{\displaystyle (l+u)}$ in
(\ref{eqnTwoViews}) with $\gamma_{2}$ and $\gamma_{3}$. Thus, the
primal problem can be reformulated as
\begin{eqnarray}
\min_{\boldsymbol{\alpha}_1,\boldsymbol{\alpha}_2,
\boldsymbol{\xi}_1, \boldsymbol{\xi}_2} && F_0=
\frac{1}{2l}\sum_{i=1}^{l}({\xi}_1^{i}+{\xi}_2^{i})+\gamma_{1}({\boldsymbol{\alpha}}_1^{\top}K_1\boldsymbol{\alpha}_1+
\nonumber \\
&&\quad
{\boldsymbol{\alpha}}_2^{\top}K_2\boldsymbol{\alpha}_2)+\gamma_{2}(\boldsymbol{\alpha}_1^{\top}K_1L_1K_1\boldsymbol{\alpha}_1+\boldsymbol{\alpha}_2^{\top}K_2L_2K_2\boldsymbol{\alpha}_2)+
\nonumber \\
&&\quad \gamma_{3} (K_1\boldsymbol{\alpha}_1-K_2\boldsymbol{\alpha}_2)^\top(K_1\boldsymbol{\alpha}_1-K_2\boldsymbol{\alpha}_2)\nonumber\\
\mbox{s.t.} && \left\{
\begin{array}{l}
y_{i}(\sum_{j=1}^{l+u}\alpha_{1}^{j}k_1(x_{j},x_{i}))\geq 1-\xi_1^{i}, \\
y_{i}(\sum_{j=1}^{l+u}\alpha_{2}^{j}k_2(x_{j},x_{i}))\geq 1-\xi_2^{i}, \\
\xi_1^{i}, \;\; \xi_2^{i}\geq 0, \quad i=1,\ldots,l\;,
\end{array}
\right. 
\label{eqn0514}
\end{eqnarray}
where $y_i \in \{+1,-1\}$, $\gamma_1, \gamma_2, \gamma_3\geq 0$.
Note that the additional bias terms  are embedded in the weight
vectors of the classifiers by using the example representation of
augmented vectors.

We present two theorems concerning the convexity and strong duality
(which means the optimal value of a primal problem is equal to that
of its Lagrange dual problem~\cite{Boydcoxbook04}) of
problem~(\ref{eqn0514}) with proofs omitted.

\begin{theorem}
Problem~(\ref{eqn0514}) is a convex optimization problem.
\end{theorem}
\begin{theorem}
Strong duality holds for problem~(\ref{eqn0514}). \label{corolla1}
\end{theorem}

Suppose $\lambda_1^i, \lambda_2^i \geq 0$ are the Lagrange
multipliers associated with the first two sets of inequality
constraints of problem (\ref{eqn0514}). Define
$\boldsymbol{\lambda}_1=(\lambda_{1}^{1},...,\lambda_{1}^{l})^\top$
and
$\boldsymbol{\lambda}_2=(\lambda_{2}^{1},...,\lambda_{2}^{l})^\top$.
It can be shown that the Lagrangian dual optimization problem with
respect to $\boldsymbol{\lambda}_1$ and $\boldsymbol{\lambda}_2$ is
a quadratic program. Classifier parameters $\boldsymbol{\alpha}_{1}$
and $\boldsymbol{\alpha}_{2}$ used by (\ref{eqn041702}) can be
solved readily after we get $\boldsymbol{\lambda}_1$ and
$\boldsymbol{\lambda}_2$.

\section{Theoretical Analysis}
\label{secTheAna}

In this section, we give a theoretical analysis of the
generalization error of the MvLapSVM method in terms of the theory
of Rademacher complexity bounds.

\subsection{Background Theory}
Some important background on Rademacher complexity theory is
introduced as follows.
\begin{definition}[Rademacher complexity, \cite{JohnbookKerMat04,Bartlett02Rade,Rosenberg07Rad}]
For a sample $S=\{x_1,\ldots,x_l\}$ generated by a distribution
$\mathcal {D}_x$ on a set $X$ and a real-valued function class
$\mathcal {F}$ with domain $X$, the empirical Rademacher complexity
of $\mathcal {F}$ is the random variable
$$\hat{R}_l(\mathcal {F})=\mathbb{E}_{\boldsymbol{\sigma}}
[\sup_{f\in\mathcal {F}}|\frac{2}{l}\sum_{i=1}^l \sigma_i
f(x_i)||x_1,\ldots,x_l],$$ where
$\boldsymbol{\sigma}=\{\sigma_1,\ldots,\sigma_l\}$ are independent
uniform $\{\pm 1\}$-valued (Rademacher) random variables. The
Rademacher complexity of $\mathcal {F}$ is
$${R}_l(\mathcal {F})=\mathbb{E}_S[\hat{R}_l(\mathcal {F})]=\mathbb{E}_{S\boldsymbol{\sigma}}
[\sup_{f\in\mathcal {F}}|\frac{2}{l}\sum_{i=1}^l \sigma_i
f(x_i)|].$$
\end{definition}

\begin{lemma}[\cite{JohnbookKerMat04}]
Fix $\delta\in (0,1)$ and let $\mathcal {F}$ be a class of functions
mapping from an input space $Z$ (for supervised learning having the
form $Z=X \times Y$) to $[0,1]$. Let $(z_i)_{i=1}^l$ be drawn
independently according to a probability distribution $\mathcal
{D}$. Then with probability at least $1-\delta$ over random draws of
samples of size $l$, every $f\in\mathcal {F}$ satisfies
\begin{eqnarray}
\mathbb{E}_{\mathcal {D}}[f(z)]&\leq& \hat{\mathbb{E}}[f(z)] +
R_l(\mathcal {F}) +\sqrt{\frac{\ln(2/\delta)}{2l}} \nonumber\\
&\leq& \hat{\mathbb{E}}[f(z)] + \hat{R}_l(\mathcal {F})
+3\sqrt{\frac{\ln(2/\delta)}{2l}} , \nonumber
\end{eqnarray}
where $\hat{\mathbb{E}}[f(z)]$ is the empirical error averaged on
the $l$ examples. \label{ThmGerErr}
\end{lemma}

Note that the above lemma is also applicable if we replace $[0, 1]$
by $[-1, 0]$. This can be justified by simply following the proof of
Lemma~\ref{ThmGerErr}, as detailed in~\cite{JohnbookKerMat04}.

\subsection{The Generalization Error of MvLapSVM}
We obtain the following theorem regarding the generalization error
of MvLapSVM, which is similar to one theorem
in~\cite{Farquhar06svm2k}. The prediction function in MvLapSVM is
adopted as the average of prediction functions from two views
\begin{equation}
g=\frac{1}{2}(f_1+f_2).
\end{equation}

\begin{theorem}
Fix $\delta\in (0,1)$ and let $\mathcal {F}$ be the class of
functions mapping from $Z=X\times Y$ to $\textbf{R}$ given by
$\tilde{f}(x,y)=-yg(x)$ where $g=\frac{1}{2}(f_1+f_2)\in\mathcal
{G}$ and $\tilde{f}\in\mathcal {F}$. Let
$S=\{(x_1,y_1),\cdots,(x_l,y_l)\}$ be drawn independently according
to a probability distribution $\mathcal {D}$. Then with probability
at least $1-\delta$ over samples of size $l$, every $g\in\mathcal
{G}$ satisfies
\begin{eqnarray}
P_{\mathcal {D}}(y\neq sgn(g(\textbf{x}))) \leq &&
\frac{1}{2l}\sum_{i=1}^{l}(\xi_1^i+\xi_2^i)+2\hat{R}_l(\mathcal
{G})+ 3\sqrt{\frac{\ln(2/\delta)}{2l}}, \nonumber
\end{eqnarray}
\label{ThmMvLapErr}
\end{theorem}
where $\xi_1^i=(1-y_i f_1({x}_i))_+$ and $\xi_2^i=(1-y_i
f_2({x}_i))_+$.
\begin{proof}
Let $H(\cdot)$ be the Heaviside function that returns 1 if its
argument is greater than 0 and zero otherwise. Then it is clear to
have
\begin{equation}
P_{\mathcal {D}}(y\neq sgn(g(\textbf{x})))=\mathbb{E}_{\mathcal
{D}}[H(-yg(\textbf{x}))]. \label{eqn07051}
\end{equation}

Consider a loss function $\mathcal {A}: \textbf{R}\rightarrow[0,1]$,
given by
$$
\mathcal{A}(a)=\left\{
\begin{array}{ll}
1, &\quad \mbox{if  $a\geq 0$};\\
1+a, &\quad \mbox{if  $-1\leq a\leq 0$}; \\
0, &\quad \mbox{otherwise}.
\end{array}
\right.
$$
By Lemma~\ref{ThmGerErr} and since function $\mathcal {A}-1$
dominates $H-1$, we have~\cite{JohnbookKerMat04}
\begin{eqnarray}
&&\mathbb{E}_{\mathcal {D}}[H(\tilde{f}(x,y))-1]\leq
\mathbb{E}_{\mathcal {D}}[\mathcal {A}(\tilde{f}(x,y))-1] \nonumber\\
&&\leq \hat{\mathbb{E}}[\mathcal {A}(\tilde{f}(x,y))-1] +
\hat{R}_l((\mathcal {A}-1)\circ\mathcal {F})
+3\sqrt{\frac{\ln(2/\delta)}{2l}} . \nonumber
\end{eqnarray}
Therefore,
\begin{eqnarray}
&&\mathbb{E}_{\mathcal {D}}[H(\tilde{f}(x,y))]\nonumber\\
\leq&& \hat{\mathbb{E}}[\mathcal {A}(\tilde{f}(x,y))] +
\hat{R}_l((\mathcal {A}-1)\circ\mathcal {F})
+3\sqrt{\frac{\ln(2/\delta)}{2l}} . \label{eqn07052}
\end{eqnarray}

In addition, we have
\begin{eqnarray}
\hat{E}[\mathcal
{A}(\tilde{f}({x},y))]&\leq&\frac{1}{l}\sum_{i=1}^{l}(1-y_i
g({x}_i))_{+}\nonumber\\
&=&\frac{1}{2l}\sum_{i=1}^{l}(1-y_i
f_1({x}_i)+1-y_i f_2({x}_i))_{+} \nonumber\\
&\leq&\frac{1}{2l}\sum_{i=1}^{l}[(1-y_i f_1({x}_i))_++(1-y_i
f_2({x}_i))_+]
\nonumber\\
&=&\frac{1}{2l}\sum_{i=1}^{l}(\xi_1^i+\xi_2^i), \label{eqn07053}
\end{eqnarray}
where $\xi_1^i$ denotes the amount by which function $f_1$ fails to
achieve margin $1$ for $(x_i,y_i)$ and $\xi_2^i$ applies similarly
to function $f_2$.

Since $(\mathcal {A}-1)(0)=0$, we can apply the Lipschitz
condition~\cite{Bartlett02Rade} of function $(\mathcal {A}-1)$ to
get
\begin{equation}
\hat{R}_l((\mathcal {A}-1)\circ \mathcal {F}) \leq
2\hat{R}_l(\mathcal {F}). \label{eqn07054}
\end{equation}
It remains to bound the empirical Rademacher complexity of the class
$\mathcal {F}$. With $y_i\in\{+1,-1\}$, we have
\begin{eqnarray}
\hat{R}_l(\mathcal {F})&=&\mathbb{E}_{\boldsymbol{\sigma}}
[\sup_{f\in\mathcal {F}}|\frac{2}{l}\sum_{i=1}^l \sigma_i
\tilde{f}(x_i,y_i)|] \nonumber\\
&=&\mathbb{E}_{\boldsymbol{\sigma}} [\sup_{g\in\mathcal
{G}}|\frac{2}{l}\sum_{i=1}^l \sigma_i y_i g(x_i)|]\nonumber\\
&=&\mathbb{E}_{\boldsymbol{\sigma}} [\sup_{g\in\mathcal
{G}}|\frac{2}{l}\sum_{i=1}^l \sigma_i g(x_i)|] =
\hat{R}_l(\mathcal{G}). \label{eqn07055}
\end{eqnarray}

Now combining (\ref{eqn07051})$\sim$(\ref{eqn07055}) reaches the
conclusion of this theorem. \qed
\end{proof}

\subsection{The Empirical Rademacher Complexity $\hat{R}_l(\mathcal
{G})$} \label{secERC}

In this section, we give the expression of $\hat{R}_l(\mathcal {G})$
used in Theorem~\ref{ThmMvLapErr}. $\hat{R}_l(\mathcal {G})$ is also
important in identifying the different roles of regularization terms
in the MvLapSVM approach. The techniques adopted to derive
$\hat{R}_l(\mathcal {G})$ is analogical to and inspired by those
used for analyzing co-RLS
in~\cite{Sindhwani08anRKHS,Rosenberg07Rad}.

The loss function $\hat{L}: \mathcal {H}^1\times\mathcal
{H}^2\rightarrow[0,\infty)$ in (\ref{eqnTwoViews}) with
$\hat{L}=\frac{1}{2l}\sum_{i=1}^l
[(1-y_{i}f_1(x_{i}))_{+}+(1-y_{i}f_2(x_{i}))_{+}]$ satisfies
\begin{equation}
\hat{L}(0,0)= 1.
\end{equation}
We now derive the regularized function class $\mathcal {G}$ from
which our predictor $g$ is drawn.

Let $Q(f_1,f_2)$ denote the objective function in
(\ref{eqnTwoViews}). Substituting the predictors $f_1\equiv 0$ and
$f_2\equiv 0$ into $Q(f_1,f_2)$ results in an upper bound
\begin{equation}
\min_{f_1,f_2 \in \mathcal {H}^1\times\mathcal {H}^2} Q(f_1,f_2)
\leq Q(0,0)=\hat{L}(0,0)= 1.
\end{equation}
Because each term in $Q(f_1,f_2)$ is nonnegative,  the optimal
function pair $(f_1^*,f_2^*)$ minimizing $Q(f_1,f_2)$ must be
contained in
\begin{eqnarray}
\label{eqn041703} \mathcal {H}&=&\{(f_1,f_2):
\gamma_1(\|f_1\|^{2}+\|f_2\|^{2})+
\gamma_{2}({\textbf{f}_{1u}}^{\top}L_{1u}{\textbf{f}_{1u}}+\nonumber\\
&&{\textbf{f}_{2u}}^{\top}L_{2u}{\textbf{f}_{2u}})+
\gamma_{3}\sum_{i=l+1}^{l+u}[f_{1}(x_{i})-f_{2}(x_{i})]^{2} \leq
1\},
\end{eqnarray}
where parameters $\gamma_1$, $\gamma_2$, $\gamma_3$ are from
(\ref{eqn0514}),
$\textbf{f}_{1u}=(f_1(x_{l+1}),...,f_1(x_{l+u}))^\top$,
$\textbf{f}_{2u}=(f_2(x_{l+1}),...,f_2(x_{l+u}))^\top$, and $L_{1u}$
and $L_{2u}$ are the unnormalized graph Laplacians for the graphs
only involving the unlabeled examples (to make theoretical analysis
on $\hat{R}_l(\mathcal {G})$ feasible, we temporarily assume that
the Laplacians in (\ref{eqnTwoViews}) are unnormalized).

The final predictor is found out from the function class
\begin{equation}
\mathcal {G}=\{{x}\rightarrow \frac{1}{2}[f_1({x})+f_2({x})]:
(f_1,f_2)\in \mathcal {H}\},
\end{equation}
which does not depend on the labeled examples.

The complexity $\hat{R}_l(\mathcal {G})$ is
\begin{equation}
\label{eqnRlGdef} \hat{R}_l(\mathcal
{G})=\mathbb{E}_{\boldsymbol{\sigma}}[\sup_{(f_1,f_2)\in \mathcal
{H}}|\frac{1}{l}\sum_{i=1}^{l}\sigma_i(f_1({x}_i)+f_2({x}_i))|].
\end{equation}

To derive the Rademacher complexity, we first convert from a
supremum over the functions  to a supremum over their corresponding
expansion coefficients. Then, the Kahane-Khintchine
inequality~\cite{LatalaKK94} is employed to bound the expectation
over $\boldsymbol{\sigma}$ above and below, and give a computable
quantity. The following theorem summarizes our derived Rademacher
complexity.

\begin{theorem}
Suppose $\mathcal {S}=K_{1l}(\gamma_1 K_1+\gamma_2K_{1u}^\top
L_{1u}K_{1u})^{-1}K_{1l}^\top+K_{2l}(\gamma_1
K_2+\gamma_2K_{2u}^\top L_{2u}K_{2u})^{-1}K_{2l}^\top$,
$\Theta=K_{1u}(\gamma_1 K_1+\gamma_2K_{1u}^\top
L_{1u}K_{1u})^{-1}K_{1u}^\top+ K_{2u}(\gamma_1
K_2+\gamma_2K_{2u}^\top L_{2u}K_{2u})^{-1}K_{2u}^\top$, $\mathcal
{J}=K_{1u}(\gamma_1 K_1+\gamma_2K_{1u}^\top
L_{1u}K_{1u})^{-1}K_{1l}^\top- K_{2u}(\gamma_1
K_2+\gamma_2K_{2u}^\top L_{2u}K_{2u})^{-1}K_{2l}^\top$, where
$K_{1l}$ and $K_{2l}$ are respectively the first $l$ rows of $K_1$
and $K_2$, and $K_{1u}$ and $K_{2u}$ are respectively the last $u$
rows of $K_1$ and $K_2$. Then we have
$\frac{U}{\sqrt{2}l}\leq\hat{R}_l(\mathcal {G})\leq\frac{U}{l}$ with
$U^2=tr(\mathcal {S})-\gamma_3 tr(\mathcal {J}^\top (I+\gamma_3
\Theta)^{-1}\mathcal {J})$.
\end{theorem}

\section{Experiments}
\label{secExp}

We performed multi-view semi-supervised learning experiments on a
synthetic and two real-world classification problems. The Laplacian
SVM (LapSVM)~\cite{Belkin06ManfReg}, co-Laplacian SVM
(CoLapSVM)~\cite{Sindhwani05CoLapSVM}, manifold co-regularization
(CoMR)~\cite{Sindhwani08anRKHS} and co-SVM (a counterpart of the
co-RLS in~\cite{Sindhwani05CoLapSVM}) are employed for comparisons
with our proposed method. For each method, besides considering the
prediction function $(f_1+f_2)/2$ for the combined view, we also
consider the prediction functions $f_1$ and $f_2$ from the separate
views.

Each data set is divided into a training set (including labeled and
unlabeled training data), a validation set and a test set. The
validation set is used to select regularization parameters from the
range $\{10^{-10}, 10^{-6}, 10^{-4}, 10^{-2}, 1, 10, 100\}$, and
choose which prediction function should be used. With the identified
regularization parameter and prediction function, performances on
the test data and unlabeled training data would be evaluated. The
above process is repeated at random for ten times, and the reported
performance is the averaged accuracy and the corresponding standard
deviation.

\subsection{Two-Moons-Two-Lines Synthetic Data}

\begin{figure}[b]
      \centering
      \subfigure[Two-moons view]{
      \includegraphics[width=2.2in]{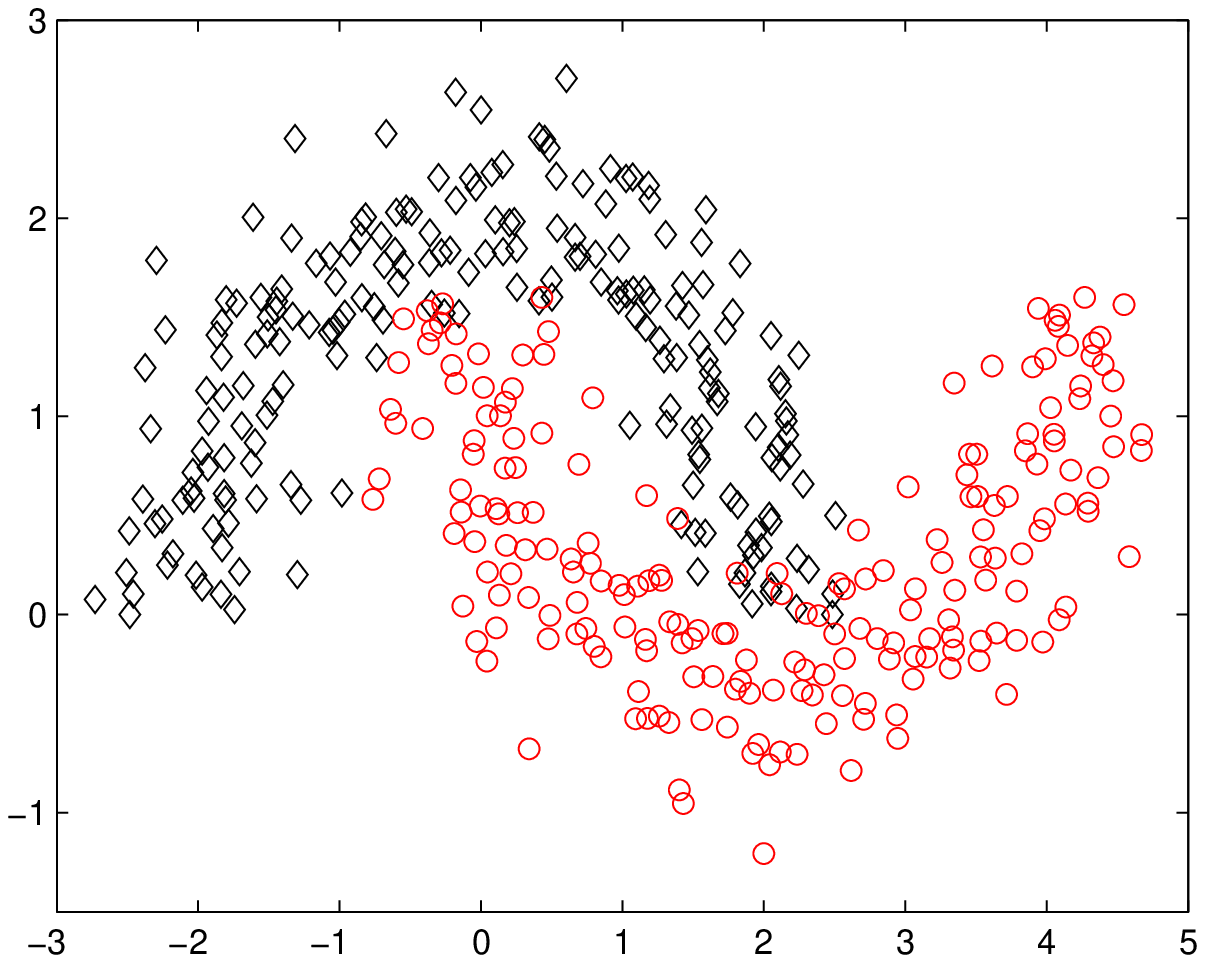}
      \label{figMvLapTMTL1_a}}
      \subfigure[Two-lines view]{
      \includegraphics[width=2.2in]{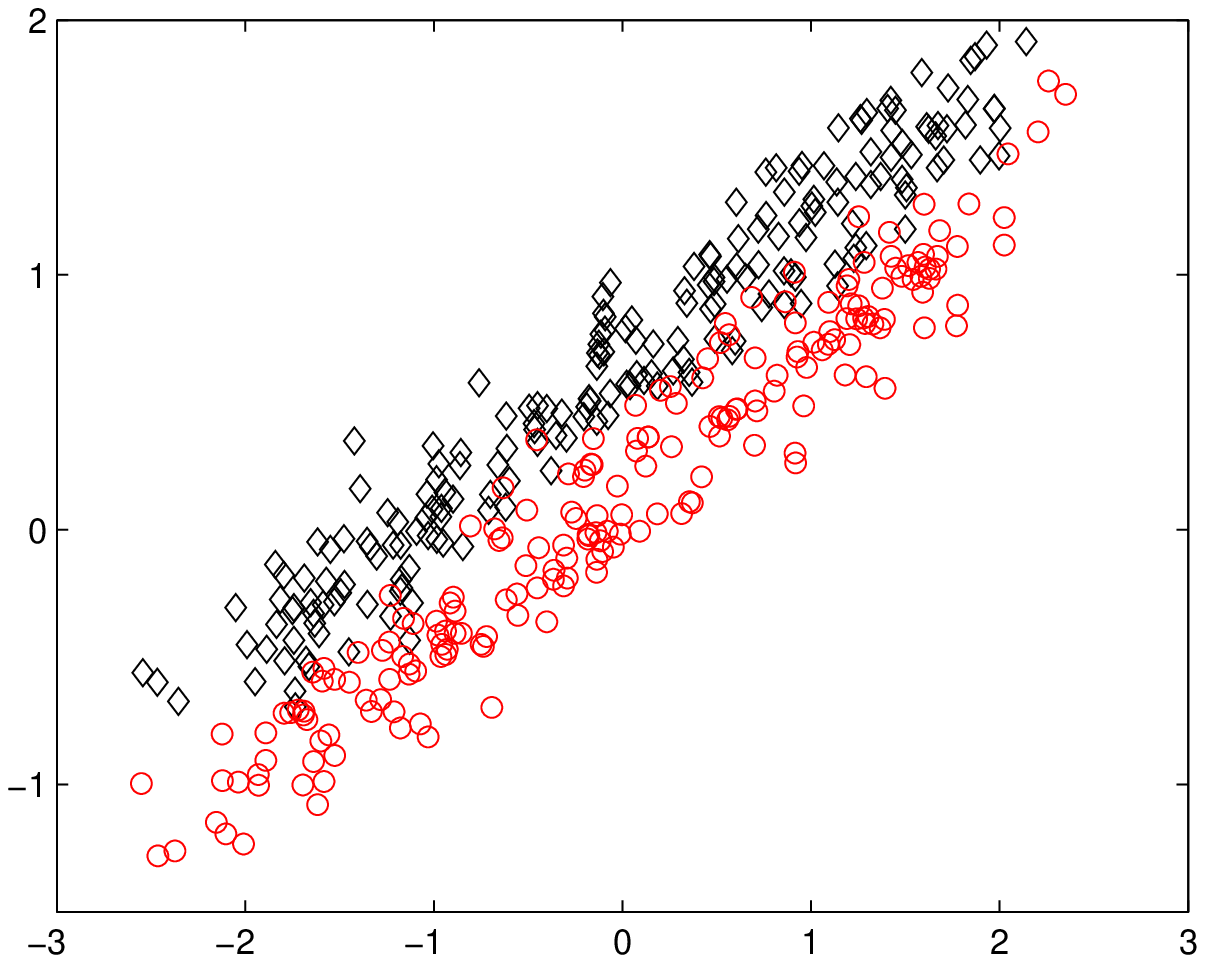}
      \label{figMvLapTMTL1_b}}
      \caption{Distribution of the two-moons-two-lines data}
      \label{figTMTL3data}
\end{figure}

This synthetic data set is generated similarly to the toy example
used in~\cite{Sindhwani05CoLapSVM}. Noisy examples in two classes
appear as two moons in one view and two parallel lines in the other
view, and points on one moon are enforced at random to associate
with points on one line (see Fig.~\ref{figTMTL3data} for an
illustration). The sizes for labeled training set, unlabeled
training set, validation set and test set are $10$, $200$, $100$ and
$100$, respectively.

\begin{table}[t]
\renewcommand{\arraystretch}{1.5}
\caption{Classification accuracies and standard deviations (\%) of
different methods on the synthetic data} \label{table_TMTL3} \vskip
0.2in \centering
\begin{tabular}{|c*{5}{|@{\hspace{0.035 in}}c@{\hspace{0.035 in}}}|}
\hline
             & LapSVM &CoLapSVM & CoMR & Co-SVM & MvLapSVM\\
\hline
$\mathbb{T}$ &91.40 (1.56)& 93.40 (3.07) &91.20 (1.60) &96.30 (1.95) &{\bf 96.90} (1.70)\\
\hline
$\mathbb{U}$ &90.60 (2.33)& 93.55 (2.72) &90.90 (2.02) &96.40 (1.61) &{\bf 96.40} (1.46)\\
\hline
\end{tabular}
\end{table}

As in~\cite{Sindhwani05CoLapSVM}, a Gaussian and linear kernel are
respectively chosen for the two-moons and two-lines view. The
classification accuracies of different methods on this data set are
shown in Table~\ref{table_TMTL3}, where $\mathbb{T}$ and $\mathbb
{U}$ means accuracies on the test data and unlabeled training data,
respectively, and best accuracies are indicated in bold (if two
methods bear the same accuracy, the smaller standard deviation will
identify the better method).

From this table, we see that methods solely integrating manifold or
multi-view regularization give good performance, which indicates the
usefulness of these regularization concerns. Moreover, among all the
methods, the proposed MvLapSVM performs best both on the test set
and unlabeled training set.

\subsection{Image-Text Classification}
\label{SecTask1}

\begin{figure}[b]
      \centering
      \includegraphics[width=4.5in]{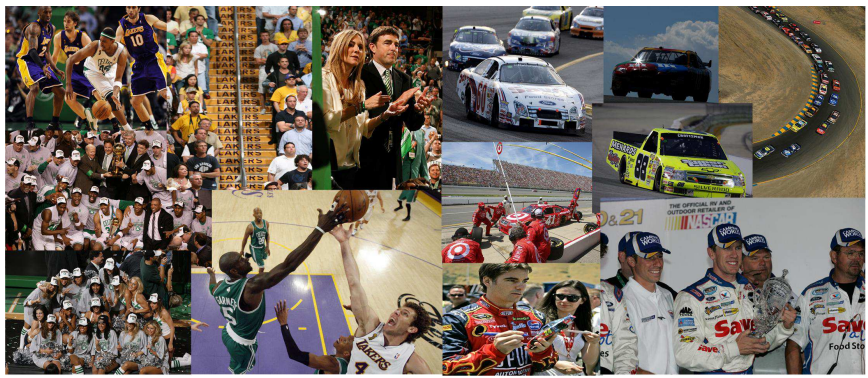}
      \caption{NBA (left) and NASCAR (right) images}
      \label{figNBA_NASCARPics}
\end{figure}

We collected this data set from the sports gallery of the yahoo!
website in 2008. It includes 420 NBA images and 420 NASCAR images,
some of which are shown in Fig.~\ref{figNBA_NASCARPics}. For each
image, there is an attached short text describing content-related
information. Therefore, image and text constitute the two views of
this data set.

Each image is normalized to be a $32\times32$-sized gray image.
Feature extraction for the texts is done by removing stop words,
punctuation and numbers and then applying Porter's
stemming~\cite{Porter80}. In addition, words that occur in five or
fewer documents were ignored. After this preprocessing, each text
has a TFIDF feature~\cite{Salton88} of 296 dimensions.

The sizes for labeled training set, unlabeled training set,
validation set and test set are $10$, $414$, $206$ and $210$,
respectively. Linear kernels are used for both views. The
performance is reported in Table~\ref{table_IT} where co-SVM ranks
first on the test set while MvLapSVM outperforms all the other
methods on the unlabeled training set. If we take the average of the
accuracies on the test set and unlabeled training set, clearly our
MvLapSVM ranks first.

\begin{table}[t]
\renewcommand{\arraystretch}{1.5}
\caption{Classification accuracies and standard deviations (\%) of
different methods on the NBA-NASCAR data} \label{table_IT} \vskip
0.2in \centering
\begin{tabular}{|c*{5}{|@{\hspace{0.035 in}}c@{\hspace{0.035 in}}}|}
\hline
             & LapSVM &CoLapSVM & CoMR & Co-SVM & MvLapSVM\\
\hline
$\mathbb{T}$ &99.33 (0.68)& 98.86 (1.32) &99.38 (0.68) &{\bf 99.43} (0.59) &99.38 (0.64)\\
\hline
$\mathbb{U}$ &99.03 (0.88)& 98.55 (0.67) &98.99 (0.90) &98.91 (0.38) &{\bf 99.54} (0.56)\\
\hline
\end{tabular}
\end{table}

\subsection{Web Page Categorization}
\label{secwebpage}

In this subsection, we consider the problem of classifying web
pages. The data set consists of 1051 two-view web pages collected
from the computer science department web sites at four U.S.
universities: Cornell, University of Washington, University of
Wisconsin, and University of Texas~\cite{Blum98Cot}. The task is to
predict whether a web page is a course home page or not. Within the
data set there are a total of 230 course home pages. The first view
of the data is the words appearing on the web page itself, whereas
the second view is the underlined words in all links pointing to the
web page from other pages. We preprocess each view according to the
feature extraction procedure used in Section~\ref{SecTask1}. This
results in 2332 and 87-dimensional vectors in view 1 and view 2
respectively~\cite{Sun08ICDMw}. Finally, document vectors were
normalized to TFIDF features.

\begin{table}
\renewcommand{\arraystretch}{1.5}
\caption{Classification accuracies and standard deviations (\%) of
different methods on the web page data} \label{table_Webpage} \vskip
0.2in \centering
\begin{tabular}{|c*{5}{|@{\hspace{0.035 in}}c@{\hspace{0.035 in}}}|}
\hline
             & LapSVM &CoLapSVM & CoMR & Co-SVM & MvLapSVM\\
\hline
$\mathbb{T}$ &94.02 (2.66)& 93.68 (2.98) &94.02 (2.24) &93.45 (3.21) &{\bf 94.25} (1.62)\\
\hline
$\mathbb{U}$ &93.33 (2.40)& 93.39 (2.44) &93.26 (2.19) &93.16 (2.68) &{\bf 93.53} (2.04)\\
\hline
\end{tabular}
\end{table}

The sizes for labeled training set, unlabeled training set,
validation set and test set are $12$, $519$, $259$ and $261$,
respectively. Linear kernels are used for both views.
Table~\ref{table_Webpage} gives the classification results obtained
by different methods. MvLapSVM outperforms all the other methods on
both the test data and unlabeled training data.

\section{Conclusion}
\label{secCon}

In this paper, we have proposed a new approach for multi-view
semi-supervised learning. This approach is an extension of SVMs for
multi-view semi-supervised learning with manifold and multi-view
regularization integrated. We have proved the convexity and strong
duality of the primal optimization problem, and used the dual
optimization to solve classifier parameters. Moreover, theoretical
results on the generalization performance of the MvLapSVM approach
and the empirical Rademacher complexity which can indicate different
roles of regularization terms have been made. Experimental practice
on multiple data sets has also manifested the effectiveness of the
proposed method.

The MvLapSVM is not a special case of the framework that Rosenberg
et al. formulated in~\cite{Rosenberg09SPM}. The main difference is
that they require the loss functional depends only on the combined
prediction function, while we use here a slightly general loss which
has a separate dependence on the prediction function from each view.
Their framework does not subsume our approach.

For future work, we mention the following three directions.
\begin{itemize}
\item {\bf Model selection:} As is common in many machine learning
algorithms, our method has several regularization parameters to set.
Usually, a held out validation set would be used to perform
parameter selection, as what was done in this paper. However, for
the currently considered semi-supervised learning, this is not very
natural because there is often a small quantity of labeled examples
available. Model selection for semi-supervised learning using no or
few labeled examples is worth further studying.
\item {\bf Multi-class classification:} The MvLapSVM algorithm implemented in
this paper is intended for binary classification. Though the usual
one-versus-rest, one-versus-another strategy, which converts a
problem from multi-class to binary classification, can be adopted
for multi-class classification, it is not optimal. Incorporating
existing ideas of multi-class SVMs~\cite{HsuMulClaSVM02} into the
MvLapSVM approach would be a further concern.
\item {\bf Regularization selection:} In this paper, although the MvLapSVM algorithm obtained good
results, it involves more regularization terms than related methods
and thus needs more assumptions. For some applications, these
assumptions might not hold. Therefore, a probably interesting
improvement could be comparing different kinds of regularizations
and attempting to select those promising ones for each application.
This also makes it possible to weight different views unequally.
\end{itemize}

\subsubsection*{Acknowledgments.} This work was supported in part by
the National Natural Science Foundation of China under Project
61075005, and the Fundamental Research Funds for the Central
Universities.


\begin{thebibliography}{99}
\bibitem{Chapelle06SSLbook}
Chapelle, O., Sch\"{o}lkopf, B., Zien, A.:  Semi-supervised
Learning. MIT Press, Cambridge, MA (2006)

\bibitem{Culp08GraBasSSL}
Culp, M., Michailidis, G.: Graph-based Semi-supervised Learning.
IEEE Transactions on Pattern Analysis and Machine Intelligence, Vol.
30 (2008) 174--179

\bibitem{Sun10JMLR}
Sun, S., Shawe-Taylor, J.: Sparse Semi-supervised Learning using
Conjugate Functions. Journal of Machine Learning Research, Vol. 11
(2010) 2423--2455

\bibitem{Zhu08tecrep}
Zhu, X.: Semi-supervised Learning Literature Survey. Technical
Report, 1530, University of Wisconsin-Madison (2008)

\bibitem{Sun11VC}
Sun, S., Jin, F., Tu, W.: View Construction for Multi-view
Semi-supervised Learning. Lecture Notes in Computer Science, Vol.
6675 (2011) 595--601

\bibitem{VapnikSLT98}
Vapnik, V.N.:  Statistical Learning Theory. Wiley, New York (1998)

\bibitem{Joachims99TSVM}
Joachims, T.:  Transductive Inference for Text Classification using
Support Vector Machines. Proceedings of the 16th International
Conference on Machine Learning (1999) 200--209

\bibitem{Bennett99S3VM}
Bennett, K., Demiriz, A.: Semi-supervised Support Vector Machines.
Advances in Neural Information Processing Systems, Vol. 11 (1999)
368--374

\bibitem{Fung01S3VM}
Fung, G.,  Mangasarian, O.L.: Semi-supervised Support Vector
Machines for Unlabeled Data Classification. Optimization Methods and
Software, Vol. 15 (2001) 29--44

\bibitem{Belkin06ManfReg}
Belkin, M.,  Niyogi,  P.,  Sindhwani,  V.: Manifold Regularization:
A Geometric Framework for Learning from Labeled and Unlabeled
Exampls. Journal of Machine Learning Research, Vol. 7 (2006)
2399--2434

\bibitem{Sindhwani05CoLapSVM}
Sindhwani, V., Niyogi, P.,  Belkin, M.: A Co-regularization Approach
to Semi-supervised Learning with Multiple Views. Proceedings of the
Workshop on Learning with Multiple Views, International Conference
on Machine Learning (2005)

\bibitem{Farquhar06svm2k}
Farquhar, J.,  Hardoon,  D., Meng,  H.,  Shawe-Taylor, J., Szedmak,
S.: Two View Learning: SVM-2K, Theory and Practice. Advances in
Neural Information Processing Systems, Vol. 18 (2006) 355--362

\bibitem{Tikhonov63reg}
Tikhonov, A.N.:  Regularization of Incorrectly Posed Problems.
Soviet Mathematics Doklady, Vol. 4 (1963) 1624--1627

\bibitem{Evgeniou00RegSVM}
Evgeniou, T., Pontil,  M., Poggio, T.: Regularization Networks and
Support Vector Machines. Advances in Computational Mathematics, Vol.
13 (2000) 1--50

\bibitem{JohnbookKerMat04}
Shawe-Taylor, J.,  Cristianini, N.: Kernel Methods for Pattern
Analysis. Cambridge University Press, Cambridge, England (2004)

\bibitem{Belkin03LapEig}
Belkin, M.,  Niyogi,  P.: Laplacian Eigenmaps for Dimensionality
Reduction and Data Representation. Neural Computation, Vol. 15
(2003) 1373--1396

\bibitem{Blum98Cot}
Blum, A.,  Mitchell,  T.: Combining Labeled and Unlabeled Data with
Co-training. Proceedings of the 11th Annual Conference on
Computational Learning Theory (1998) 92--100

\bibitem{Aronszajn50}
Aronszajn, N.:  Theory of Reproducing Kernels. Transactions of the
American Mathematical Society, Vol.  68 (1950) 337--404

\bibitem{Sindhwani08anRKHS}
Sindhwani, V., Rosenberg, D.:  An RKHS for Multi-view Learning and
Manifold Co-regularization. Proceedings of the 25th International
Conference on Machine Learning (2008) 976--983

\bibitem{Kimeldorf71}
Kimeldorf, G., Wahba, G.:  Some Results on Tchebycheffian Spline
Functions. Journal of Mathematical Analysis and Applications, Vol.
33 (1971) 82--95

\bibitem{RosenbergDiss08}
Rosenberg, D.:  Semi-Supervised Learning with Multiple Views. PhD
dissertation, Department of Statistics, University of California,
Berkeley (2008)

\bibitem{Boydcoxbook04}
Boyd, S.,  Vandenberghe,  L.: Convex Optimization.  Cambridge
University Press, Cambridge, England (2004)

\bibitem{Bartlett02Rade}
Bartlett, P., Mendelson, S.:  Rademacher and Gaussian Complexities:
Risk Bounds and Structural Results. Journal of Machine Learning
Research, Vol. 3 (2002) 463--482

\bibitem{Rosenberg07Rad}
Rosenberg, D., Bartlett, P.: The Rademacher Complexity of
Co-regularized Kernel Classes. Proceedings of the 11th International
Conference on Artificial Intelligence and Statistics (2007) 396--403

\bibitem{LatalaKK94}
Latala, R.,  Oleszkiewicz, K.: On the  Best Constant in the
Khintchine-Kahane Inequality. Studia Mathematica, Vol. 109 (1994)
101--104

\bibitem{Porter80}
Porter, M.F.:  An Algorithm for Suffix Stripping. Program, Vol.  14
(1980) 130--137

\bibitem{Salton88}
Salton, G., Buckley, C.: Term-Weighting Approaches in Automatic Text
Retrieval. Information Processing and Management, Vol.  24 (1988)
513--523

\bibitem{Sun08ICDMw}
Sun, S.:  Semantic Features for Multi-view Semi-supervised and
Active Learning of Text Classification. Proceedings of the IEEE
International Conference on Data Mining Workshops (2008) 731--735

\bibitem{Rosenberg09SPM}
Rosenberg, D., Sindhwani, V., Bartlett, P.,  Niyogi, P.: Multiview
Point Cloud Kernels for Semisupervised Learning. IEEE Signal
Processing Magazine (2009) 145--150

\bibitem{HsuMulClaSVM02}
Hsu, C.W.,  Lin, C.J.: A Comparison of Methods for Multiclass
Support Vector Machines. IEEE Transactions on Neural Networks, Vol.
13 (2002) 415--425

\end{thebibliography}
\end{document}